\documentclass{article}

\PassOptionsToPackage{numbers, compress}{natbib}

\usepackage[final]{neurips_2024}

\usepackage[utf8]{inputenc} 
\usepackage[T1]{fontenc}    
\usepackage{hyperref}       
\usepackage{url}            
\usepackage{booktabs}       
\usepackage{amsfonts}       
\usepackage{nicefrac}       
\usepackage{microtype}      
\usepackage{xcolor}         
\usepackage{graphicx}
\usepackage{subfigure}
\usepackage{amsmath} 
\usepackage{float}    
\usepackage{array}

\usepackage{siunitx}    
\usepackage{adjustbox}  

\title{
    Mining the Long Tail: A Comparative Study of Data-Centric Criticality Metrics for Robust Offline Reinforcement Learning in Autonomous Motion Planning
  }

\author{%
  Antonio Guillen-Perez  \\
  Independent Researcher \\
  \texttt{antonio\_algaida@hotmail.com} \\
  \href{https://antonioalgaida.github.io/}{antonioalgaida.github.io}
}

\begin{document}

\maketitle

\begin{abstract}
Offline Reinforcement Learning (RL) presents a promising paradigm for training autonomous vehicle (AV) planning policies from large-scale, real-world driving logs. However, the extreme data imbalance in these logs, where mundane scenarios vastly outnumber rare ``long-tail'' events, leads to brittle and unsafe policies when using standard uniform data sampling. In this work, we address this challenge through a systematic, large-scale comparative study of data curation strategies designed to focus the learning process on information-rich samples. We investigate six distinct criticality weighting schemes which are categorized into three families: heuristic-based, uncertainty-based, and behavior-based. These are evaluated at two temporal scales, the individual timestep and the complete scenario. We train seven goal-conditioned Conservative Q-Learning (CQL) agents with a state-of-the-art, attention-based architecture and evaluate them in the high-fidelity Waymax simulator. Our results demonstrate that all data curation methods significantly outperform the baseline. Notably, data-driven curation using model uncertainty as a signal achieves the most significant safety improvements, reducing the collision rate by nearly three-fold (from 16.0\% to 5.5\%). Furthermore, we identify a clear trade-off where timestep-level weighting excels at reactive safety while scenario-level weighting improves long-horizon planning. Our work provides a comprehensive framework for data curation in Offline RL and underscores that intelligent, non-uniform sampling is a critical component for building safe and reliable autonomous agents.
\end{abstract}

%
\section{Introduction}
\label{sec:introduction}

The development of safe and robust autonomous vehicles (AVs) promises to be one of the most transformative technologies of our time, with the potential to dramatically reduce traffic fatalities, improve transportation efficiency, and redefine mobility. Central to this endeavor is the motion planning module, the system's core decision-making component responsible for navigating dynamic, complex, and interactive environments. While significant progress has been made, achieving human-level robustness, particularly in the face of rare and unexpected "long-tail" events, remains a primary challenge to widespread deployment.

In response to this challenge, the field has increasingly shifted from traditional, hand-engineered systems towards data-driven methods that can learn from the vast scale of real-world driving data. Offline Reinforcement Learning (Offline RL) has emerged as a particularly promising paradigm. By enabling agents to learn effective policies directly from massive, pre-existing logs of expert driving behavior, Offline RL bypasses the need for costly and potentially unsafe online exploration. This makes it exceptionally well-suited for safety-critical domains like autonomous driving, allowing us to leverage petabyte-scale datasets collected by real-world fleets. Our own prior work in structured multi-agent environments has demonstrated the power of RL for learning complex coordination policies, such as signal-free intersection management~\cite{Antonio2022Apr}, motivating its application to the more unstructured problem of general motion planning.

However, the efficacy of Offline RL is fundamentally constrained by the statistical properties of the data it learns from. Real-world driving logs are massively imbalanced; the vast majority of the data consists of mundane events like following a lane on an empty road. Critical events that require true driving intelligence (such as emergency braking to avoid a cut-in, proactively yielding to a rule-breaking agent, or executing a complex merge in dense traffic) are, by their nature, statistically rare. An agent trained with standard uniform data sampling will perform countless gradient updates on common events but will be critically underexposed to the rare events where safety is most at risk. This leads to policies that perform well on average metrics but are brittle and dangerously incompetent when faced with the long-tail scenarios that define real-world driving.

In this work, we argue for a data-centric solution to this long-tail challenge. We posit that the key to robust policy learning lies not just in the algorithm, but in intelligently curating the data distribution it sees. We present a large-scale, systematic comparative study of data curation strategies, framed as "criticality metrics," designed to identify and amplify the learning signal from the most information-rich samples. Our primary contribution is a rigorous investigation into three distinct philosophies for defining criticality:

\begin{enumerate}
    \item \textbf{Heuristic-Based}, using explicit domain knowledge;
    \item \textbf{Uncertainty-Based}, using a model's own confusion as a data-driven signal of difficulty;
    \item \textbf{Behavior-Based}, using the statistical rarity of the expert's own actions.
\end{enumerate}

We evaluate these strategies at two temporal scales (the individual timestep and the complete scenario) to understand their impact on both reactive safety and long-horizon planning. Our results provide strong evidence that moving beyond uniform sampling is essential, and that data-driven curation is a powerful and necessary tool for building the next generation of safe and reliable autonomous agents.

The complete source code, including all scripts for data processing, training, and evaluation, as well as the final trained model weights, are made publicly available in our GitHub repository: \url{https://github.com/AntonioAlgaida/LongTailOfflineRL}.


\section{Related Work}
\label{sec:related_work}

Our research is situated at the intersection of three key domains: Offline Reinforcement Learning, data-driven behavior modeling for autonomous driving, and data curation for long-tail events.

\textbf{Offline Reinforcement Learning.} The primary challenge in Offline RL is learning a robust policy from a fixed dataset without the ability to explore. This requires overcoming the distributional shift problem, where a learned policy may select out-of-distribution (OOD) actions for which the value function is erroneously high. Early approaches, such as Batch-Constrained Deep Q-Learning (BCQ)~\cite{Fujimoto2019}, addressed this by training a generative model to constrain the policy's output to actions that are similar to those in the dataset. Our work is built upon a more recent and flexible paradigm, Conservative Q-Learning (CQL)~\cite{Kumar2020}, which instead learns a conservative Q-function by adding a regularizer that explicitly minimizes the values of likely OOD actions. While other powerful methods like Implicit Q-Learning (IQL)~\cite{kostrikov2021offline} and sequence modeling approaches like the Decision Transformer~\cite{chen2021decision} have emerged, we chose CQL for its strong theoretical underpinnings and its proven effectiveness in complex, real-world domains. To further stabilize our actor, we incorporate a Behavioral Cloning (BC) auxiliary loss, a technique shown to be highly effective in minimalist approaches like TD3+BC~\cite{Fujimoto2021Jun}.

\textbf{Imitation Learning in Autonomous Driving.} Data-driven approaches for vehicle control have a long history, beginning with landmark end-to-end imitation learning (IL) systems that demonstrated the feasibility of learning steering commands directly from raw sensor data~\cite{Bojarski2016Apr}. While powerful, these Behavioral Cloning (BC) methods are known to suffer from causal confusion and compounding errors, where small initial mistakes can lead the agent into out-of-distribution states, resulting in catastrophic failures. A significant body of research has focused on mitigating these issues. One effective strategy is to augment the training process with corrective feedback, as seen in systems like ChauffeurNet, which synthesizes perturbed scenarios~\cite{Bansal2018Dec}. Our own prior work explored an alternative approach, Learning from Oracle Demonstrations (LfOD), where a learned expert model provides on-demand, corrective demonstrations to an RL agent, significantly accelerating the training process and improving policy stability~\cite{Guillen-Perez2021LfOD, Antonio2022Apr}. However, even with these advances, a fundamental performance gap can remain. In a direct comparative study, we have previously shown that even a sophisticated, Transformer-based BC policy, while achieving low imitation loss, can be brittle in long-horizon simulations compared to an Offline RL agent trained on the exact same data~\cite{Guillen-Perez2025Aug}. This prior finding that pure imitation can be insufficient for achieving true robustness provides a strong motivation for our current work's focus on enhancing the safety and reliability of Offline RL agents through advanced data curation.

\textbf{Data Curation and the Long-Tail Problem.} The quality and composition of the dataset are known to be critical for the success of Offline RL, as formalized by benchmarks like D4RL~\cite{Fu2020Apr}. Our work extends this inquiry by proposing that even within a high-quality expert dataset like the Waymo Open Motion Dataset (WOMD)~\cite{Ettinger2021}, non-uniform sampling is essential for learning from the long tail. The core principle of focusing on high-information samples was powerfully demonstrated in online RL by Prioritized Experience Replay~\cite{Schaul2015Nov}, which prioritizes transitions with high temporal-difference error. Our work adapts this philosophy to the offline setting, where the signal of "importance" must be derived from the data itself. While prior work in the AV domain has successfully used heuristics to mine for critical events~\cite{potter2024long}, our contribution is a large-scale, systematic comparison of these heuristics against novel, data-driven criticality metrics. Our uncertainty-based scoring is grounded in the well-established technique of using deep ensembles for predictive uncertainty estimation~\cite{Lakshminarayanan2016Dec}, adapting it from a static prediction setting to a dynamic policy learning context. By framing our methods as a form of automatic curriculum generation~\cite{bengio2009curriculum}, we provide a comprehensive analysis of what constitutes an "important" sample for training robust, real-world driving policies.

A critical aspect of deploying real-world agents is ensuring robustness to environmental uncertainties. This includes not only the stochastic behavior of other agents but also imperfections in the system itself, such as communication latency in connected vehicle networks. Prior work has shown that explicitly modeling and predicting these uncertainties can be critical for achieving safe, collision-free control policies~\cite{Antonio2022Mar}. While our current work focuses on data uncertainty, the principle of identifying and focusing on challenging states is a shared theme.


\section{Methodology}
\label{sec:methodology}
Our approach to tackling the long-tail problem in Offline RL is rooted in a data-centric philosophy. We augment the standard Conservative Q-Learning (CQL)~\cite{Kumar2020} algorithm with a structured data curation pipeline. The core of our methodology is the concept of a "criticality score," a signal used to guide a non-uniform data sampling process, forcing the agent to focus on the most information-rich transitions. We investigate three distinct, orthogonal philosophies for defining this criticality.

\subsection{Criticality Metrics for Data Curation}
\label{sec:crit_metrics}

The central hypothesis of our work is that non-uniform data sampling can significantly improve the safety and robustness of Offline RL agents. To test this, we introduce the concept of a "criticality score," a value assigned to each data point to guide the sampling process. We designed and compared six distinct scoring functions, categorized into three families based on the source of the criticality signal. Each family was implemented at two temporal scales: the \textbf{timestep level}, to focus on reactive safety, and the \textbf{scenario level}, to focus on long-horizon planning.

\subsubsection{Heuristic-Based Criticality (Domain Knowledge)}
This family of metrics leverages explicit, human-defined domain knowledge to identify situations that are physically or socially complex. The goal is to test if we can improve performance by directly encoding our understanding of ``what is dangerous'' into the data sampling process. We designed a suite of five distinct heuristic scores that are calculated for every timestep.

\begin{itemize}
    \item \textbf{Kinematic Volatility:} To capture rapid changes in the expert's plan, we compute the SDC's longitudinal \textit{jerk} (the rate of change of acceleration) and its \textit{yaw acceleration} (the rate of change of turn rate). High values in either of these third-order derivatives are strong indicators of a reactive maneuver, such as an emergency brake or an evasive swerve. The final score is the maximum of the normalized jerk and yaw acceleration.

    \item \textbf{Interaction Score:} To identify latent collision risks beyond simple proximity, we calculate a score based on converging trajectories. For each other agent, we compute the dot product of its relative position vector and its relative velocity vector with respect to the SDC. A large negative value indicates that two agents are on a direct collision course, even if they are currently far apart. The final score for a timestep is the maximum convergence risk posed by any single agent in the scene.

    \item \textbf{Off-Road Proximity:} This metric quantifies the immediate risk of leaving the drivable area. We compute the minimum distance from any of the SDC's four bounding box corners to the nearest physical road boundary (i.e., a \texttt{RoadEdge} feature from the map data). The score is inversely proportional to this distance, creating a high-urgency signal as the vehicle approaches a curb or median.

    \item \textbf{Lane Deviation:} To identify complex maneuvers like lane changes and sharp turns, we calculate the SDC's lateral distance from the nearest lane centerline. Unlike the off-road score, this metric captures intentional, non-nominal driving behavior that occurs within the bounds of the road.

    \item \textbf{Social Density:} As a general proxy for cognitive load, we include a simple score proportional to the number of valid agents present in the scene at each timestep.
\end{itemize}

\noindent
These five raw scores are saved individually for analysis. For training the timestep-level agent, they are combined into a final criticality score using a tuned, weighted sum.

\textbf{Scenario-Level Aggregation (`CQL-HS`).} To capture sustained difficulty, we aggregate the timestep-level scores into a single value for each scenario. We use aggregation functions that are sensitive to the nature of each signal: the 99th percentile is used for the reactive, peak-based scores (Volatility, Interaction, Off-Road Proximity) to capture the single most critical moment of the event. The standard deviation is used for Lane Deviation to best capture the variability of a maneuver. The mean is used for Social Density to measure the average complexity of the scene.

\subsubsection{Uncertainty-Based Criticality (Model Disagreement)}
This family of metrics tests a more advanced, data-driven hypothesis: the most informative samples for a powerful agent to learn from are those where a committee of simpler models is most uncertain. High model disagreement serves as a proxy for states that are ambiguous, out-of-distribution, or require complex reasoning that is not easily captured by simple imitation. This allows the system to discover its own ``known unknowns'' and create a self-generated curriculum of increasing difficulty.

To generate this uncertainty signal, we first train an ensemble of \textit{scout models}. These are lightweight Behavioral Cloning agents that share the same powerful, goal-conditioned, attention-based architecture as our final CQL agent. We train an ensemble of $K=5$ such models using K-fold cross-validation, where each scout is trained on a different 80\% subset of the training data. This process ensures that each scout learns a slightly different policy, making their disagreement a meaningful measure of epistemic uncertainty.

\begin{itemize}
    \item \textbf{Timestep (`CQL-E`):} The criticality score for an individual timestep is the \textbf{trace of the covariance matrix} of the actions predicted by the $K$ models in the ensemble. This is calculated as the sum of the variances for each action dimension (acceleration and yaw rate). A high variance signifies high model disagreement for that specific state. The resulting scores are then normalized based on the 99th percentile across the dataset to create a robust signal.

    \item \textbf{Scenario (`CQL-ES`):} A scenario's difficulty score is the \textbf{99th percentile} of its constituent timestep disagreement scores. By using a high percentile instead of a simple mean, this metric is highly sensitive to scenarios that contain at least one moment of extreme model confusion, which often correspond to the most challenging and decisive points in a complex event.
\end{itemize}

\subsubsection{Behavior-Based Criticality (Action Rarity)}
This family of metrics is designed to directly address the severe data imbalance inherent in the expert's action distribution. The underlying hypothesis is that the most informative transitions are those where the expert driver performed a statistically rare maneuver, as these actions are often necessary responses to unusual or safety-critical situations. This approach is orthogonal to our other metrics, as it is agnostic to the environmental context or model uncertainty, focusing solely on the expert's behavior.

To implement this, we first build a non-parametric model of the expert's action distribution. We process the entire training set of millions of transitions and construct a fine-grained 2D histogram over the action space (longitudinal acceleration and yaw rate). To capture the high density of near-zero actions without losing resolution on rare, high-magnitude actions, we use a non-uniform binning strategy with higher resolution near the origin.

\begin{itemize}
    \item \textbf{Timestep (`CQL-AR`):} The criticality score for an individual timestep is derived directly from the action distribution model. We calculate the \textbf{smoothed inverse frequency} of the expert's action at that timestep. This computation effectively assigns a high score to actions that fall into low-density (rare) bins of the histogram, such as emergency braking or sharp, evasive turns, while assigning a low score to the overwhelmingly common action of driving straight.

    \item \textbf{Scenario (`CQL-ARS`):} To identify entire events that required an unusual response, we aggregate the rarity signal to the scenario level. A scenario's difficulty score is the \textbf{95th percentile} of its constituent timestep action-rarity scores. This metric is highly sensitive to scenarios containing at least one peak rare maneuver, allowing the agent to learn not just the action itself but the full temporal context in which such a critical action was necessary.
\end{itemize}

To summarize our multi-faceted approach, Table~\ref{tab:methodology_summary} provides a consolidated overview of the six distinct criticality scoring methods used in our experiments.

\begin{table*}
\centering
\caption{Summary of the six data curation strategies investigated. The methods are organized by their core philosophy (the source of the criticality signal) and their temporal scope (the level at which the score is applied).}
\label{tab:methodology_summary}
\begin{adjustbox}{width=\textwidth}
\begin{tabular}{@{}llll@{}}
\toprule
\textbf{Curation Philosophy} & \textbf{Agent ID} & \textbf{Core Signal / Metric} & \textbf{Implementation Method} \\
\midrule
\multicolumn{4}{l}{\textit{\textbf{Heuristic-Based (Domain Knowledge)}}} \\
\quad Timestep-Level & \texttt{CQL-H} & Instantaneous Danger \& Complexity & Weighted sum of jerk, interaction score, off-road proximity, etc. \\
\quad Scenario-Level & \texttt{CQL-HS} & Sustained Event Difficulty & Aggregated heuristics (P99 of peak events, Std Dev of maneuvers). \\
\midrule
\multicolumn{4}{l}{\textit{\textbf{Uncertainty-Based (Model Disagreement)}}} \\
\quad Timestep-Level & \texttt{CQL-E} & State Ambiguity / Confusion & Trace of the covariance matrix of a K-fold scout model ensemble's action predictions. \\
\quad Scenario-Level & \texttt{CQL-ES} & Sustained Scenario Ambiguity & 99th percentile of the timestep-level ensemble disagreement scores. \\
\midrule
\multicolumn{4}{l}{\textit{\textbf{Behavior-Based (Action Rarity)}}} \\
\quad Timestep-Level & \texttt{CQL-AR} & Statistical Rarity of Expert Action & Smoothed inverse frequency from a global 2D action histogram. \\
\quad Scenario-Level & \texttt{CQL-ARS} & Presence of a Rare Maneuver & 95th percentile of the timestep-level action rarity scores. \\
\bottomrule
\end{tabular}
\end{adjustbox}
\end{table*}

\begin{figure}
    \centering
    \includegraphics[width=0.4\textwidth]{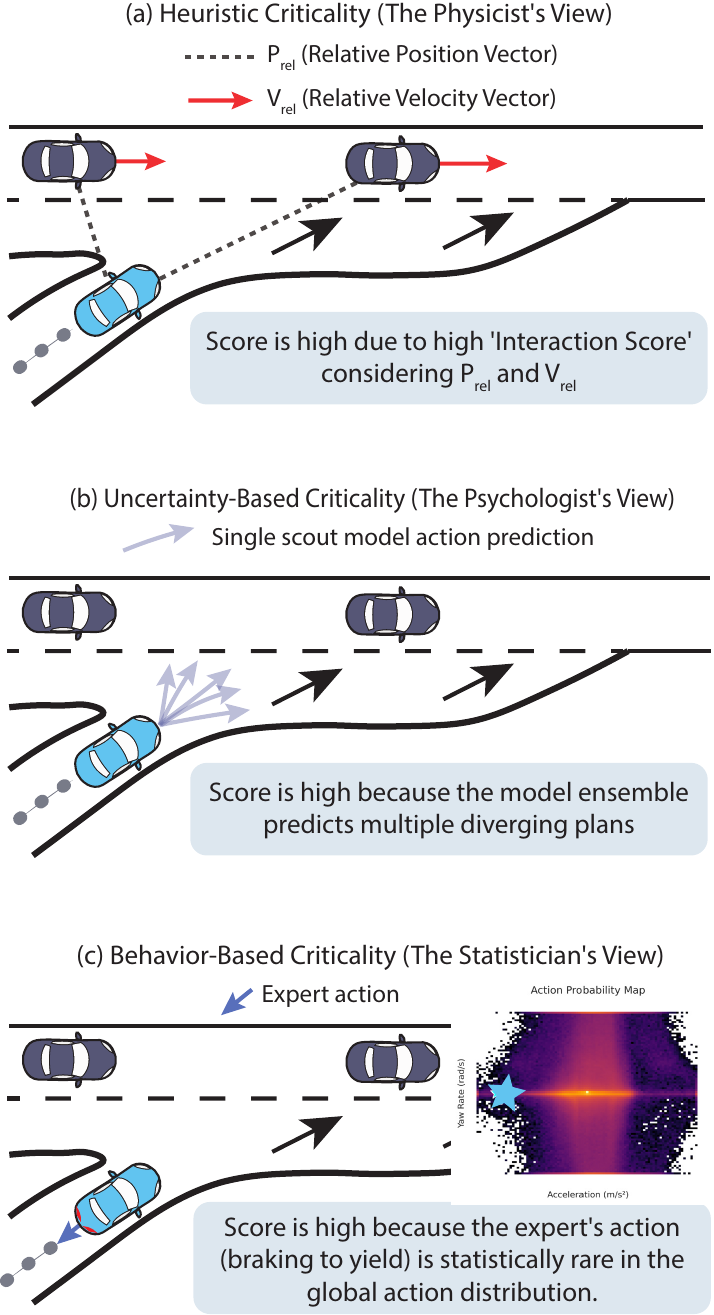} 
    \caption{\textbf{Visualizing the Three Philosophies of Criticality Scoring.} We investigate three orthogonal signals for identifying information-rich data. 
        \textbf{(a) Heuristic-Based:} Criticality is defined by physical rules. Here, the score is high because the relative position and velocity vectors of the two agents indicate a high rate of convergence (a latent collision risk).
        \textbf{(b) Uncertainty-Based:} Criticality is defined by model confusion. The score is high because an ensemble of simpler "scout" models cannot agree on a single best action, predicting a diverse set of future trajectories.
        \textbf{(c) Behavior-Based:} Criticality is defined by statistical rarity. The score is high because the expert's action (in this case, braking to yield during a merge) falls into a dark, low-density region of the global action distribution heatmap. The heatmap, derived from the entire training set, is dominated by a bright hotspot at the origin representing the overwhelmingly common action of driving straight with near-zero acceleration and steering.}
    \label{fig:crit_metrics_viz}
\end{figure}

Our three families of criticality metrics are designed to capture distinct aspects of what makes a driving event informative. We visualize these core philosophies in Figure~\ref{fig:crit_metrics_viz}.

\section{Experimental Setup}
\label{sec:exp_setup}

Our experiments are designed to provide a rigorous and fair comparison of the different data curation strategies. All agents are trained and evaluated using a unified pipeline built upon publicly available datasets and simulation tools.

\subsection{Dataset and Simulation Environment}

\textbf{Dataset.} We use the Waymo Open Motion Dataset (WOMD) v1.2~\cite{Ettinger2021}, a large-scale collection of real-world, multi-agent driving scenarios. Our experiments utilize over 100,000 scenarios from the training set for training and 1,000 held-out scenarios from the validation set for final evaluation.

\textbf{Preprocessing Pipeline.} We employ a two-stage offline data processing pipeline to convert the raw WOMD data into a format suitable for our models.
\begin{enumerate}
    \item \textbf{Parsing:} Raw scenario protobufs are parsed into an intermediate \texttt{.npz} format. This step extracts all agent trajectories, map features, and traffic light states into simple NumPy arrays, decoupling our pipeline from the source library's dependencies. We also extract the full road graph topology, including lane connectivity, and pre-computed agent roles (e.g., `is\_track\_to\_predict`).
    
    \item \textbf{Featurization:} A second script processes the \texttt{.npz} files to create the final training samples. For each valid timestep, we use a \texttt{FeatureExtractor} to generate a structured, goal-conditioned state dictionary. Crucially, the ground-truth expert actions are calculated at this stage using the official Waymax \texttt{InvertibleBicycleModel}. This ensures perfect consistency between the expert actions used for training and the dynamics model used for evaluation. The final samples, each a dictionary containing the structured state and the expert action, are saved as \texttt{.pt} files.
\end{enumerate}

\textbf{Simulator.} All agents are evaluated in closed-loop simulation using the Waymax simulator~\cite{waymo_waymax}. Waymax provides a JAX-based environment that uses the same underlying map and agent data as the WOMD, ensuring a high-fidelity evaluation.

\subsection{State and Action Representation}
\label{sec:state_action}

To enable our agent to make informed decisions, we designed a rich, structured state representation that captures the full context of the driving scene from an ego-centric perspective. The action space is defined by low-level, continuous kinematic commands.

\subsubsection{Action Space}
The output of our policy network is a 2-dimensional continuous action vector, $\mathbf{a} = [a_{\text{accel}}, a_{\text{steer}}]$, representing the desired longitudinal acceleration (in m/s²) and steering angle (in radians). For training, the ground-truth expert actions are extracted from the logged WOMD trajectories using the official Waymax \texttt{InvertibleBicycleModel}. This ensures perfect physical consistency between the training data and the simulation dynamics. The actor network's output is squashed by a Tanh activation and then rescaled to the vehicle's physical limits to ensure all commanded actions are feasible.

\subsubsection{State Space}
At each timestep, the environment state is featurized into a structured dictionary of tensors, which are then processed by entity-specific encoders in our network architecture. All spatial features are transformed into the SDC's ego-centric frame, where the SDC is at the origin `(0,0)` and oriented along the positive x-axis. This ego-centric representation is critical for policy generalization. The state is composed of the entities detailed in Table~\ref{tab:state_features}.

\begin{table}
\centering
\caption{Structured state representation provided to the agent at each timestep.}
\label{tab:state_features}
\begin{adjustbox}{width=\columnwidth} 
\begin{tabular}{@{}lll@{}}
\toprule
\textbf{Entity} & \textbf{Shape} & \textbf{Feature Description} \\
\midrule
\texttt{ego} & $(1,)$ & SDC's current speed (m/s). \\
\midrule
\texttt{agents} & $(N, 10)$ & N nearest other agents, padded. Features include: \\
& & \quad - Relative position $(x, y)$ in meters. \\
& & \quad - Relative velocity $(v_x, v_y)$ in m/s. \\
& & \quad - Heading represented as $(\cos(\theta), \sin(\theta))$. \\
& & \quad - Bounding box size (length, width) in meters. \\
& & \quad - One-hot type: $[is\_vehicle, is\_ped\_or\_cyclist]$. \\
\midrule
\texttt{map} & $(M, P \times 2)$ & M nearest lane centerlines, padded. Each is a polyline \\
& & resampled to P points, with features being the flattened \\
& & ego-centric $(x, y)$ coordinates of these points. \\
\midrule
\texttt{traffic\_lights} & $(2,)$ & State of the most relevant traffic light ahead: \\
& & \quad - $[is\_red\_ahead, \text{distance\_to\_stop\_line}]$. \\
\midrule
\texttt{goal} & $(G, 2)$ & G future waypoints from the expert's route, representing \\
& & the intended path. Features are the ego-centric $(x, y)$ \\
& & coordinates for each future waypoint. \\
\bottomrule
\end{tabular}
\end{adjustbox}
\footnotesize{\textit{N: num\_agents (16), M: num\_map\_polylines (64), P: map\_points\_per\_polyline (10), G: num\_goal\_points (5).}}
\end{table}

\subsection{Agent Architecture}
\label{sec:architecture}

\begin{figure}
    \centering
    \includegraphics[width=0.9\textwidth]{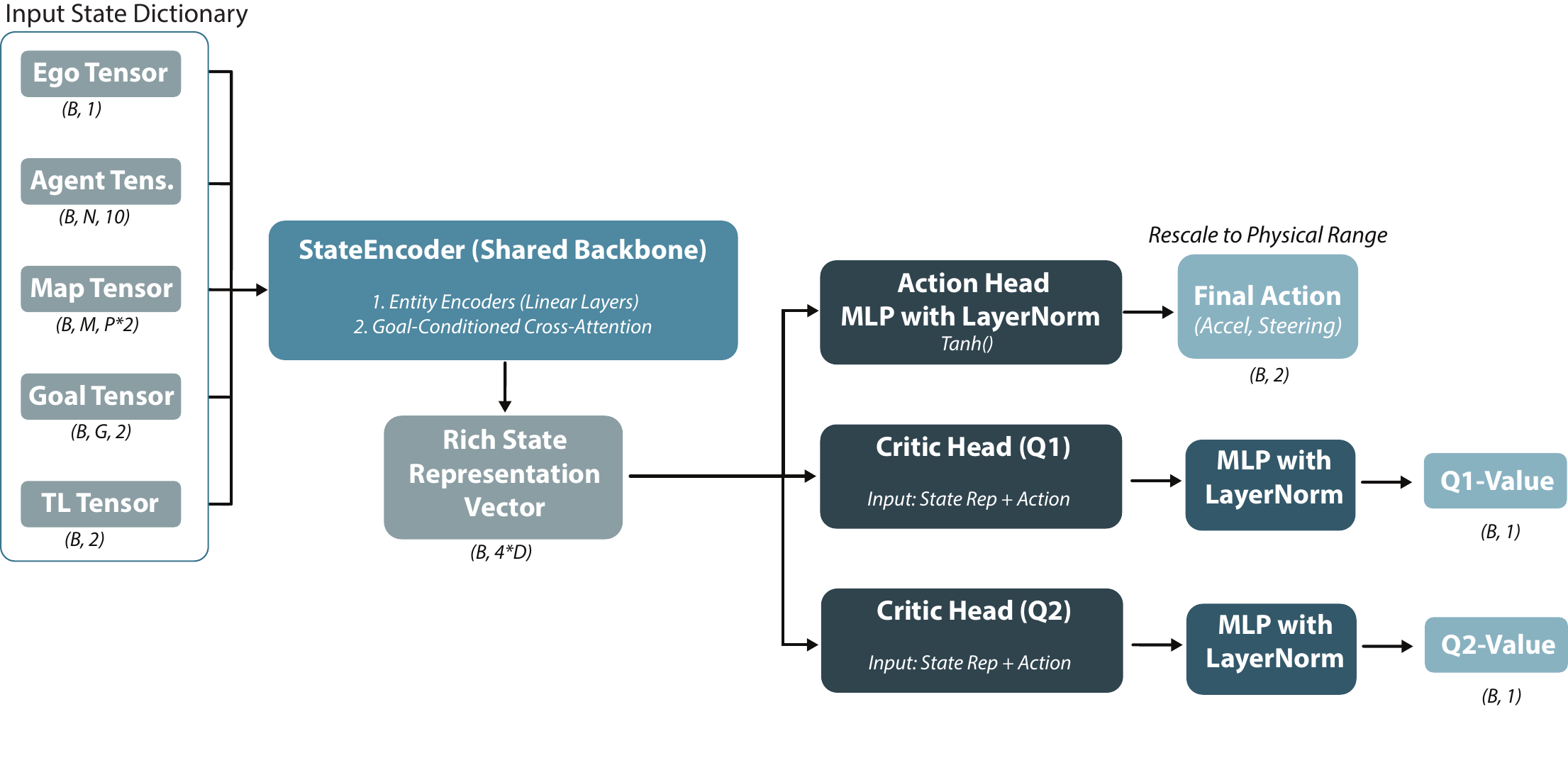}
    \caption{
        \textbf{Overview of the Goal-Conditioned, Shared-Encoder Actor-Critic Architecture.}
        The input to our model is a structured dictionary of tensors representing the scene.
        A shared \texttt{StateEncoder} backbone first processes this data.
        \textbf{(1)} Each entity type (ego, agents, map, goal, etc.) is projected into a common embedding space using dedicated linear encoders.
        \textbf{(2)} A goal-conditioned query, formed by combining the ego and goal embeddings, is used in a cross-attention mechanism to produce a context vector summarizing the most salient agents and map features.
        \textbf{(3)} The final, rich state representation is a concatenation of the ego, goal, traffic light, and attention-context embeddings.
        This shared representation is then fed into two separate heads: an \textbf{Actor head} which outputs a 2D action vector, and two independent \textbf{Critic heads} which take both the state representation and an action to produce a scalar Q-value.
    }
    \label{fig:architecture_diag}
\end{figure}

To process the structured state representation detailed in Section~\ref{sec:state_action}, we designed a goal-conditioned, attention-based actor-critic architecture. A key feature of our design is a shared \texttt{StateEncoder} backbone, which creates a rich state representation for both the actor and critic, significantly improving parameter efficiency. The complete architecture is visualized in Figure~\ref{fig:architecture_diag}.

\subsubsection{Shared StateEncoder Backbone}
The \texttt{StateEncoder} serves as a powerful perception module that reasons about the relationships between entities in the scene. Its forward pass consists of three stages:

\begin{enumerate}
    \item \textbf{Entity Encoding:} Each tensor in the input state dictionary is passed through a dedicated linear encoder with a ReLU activation. This projects the heterogeneous raw features (e.g., a 1D ego speed, a 10D agent vector) into a common, high-dimensional embedding space (we use an \texttt{embed\_dim} of 96).

    \item \textbf{Goal-Conditioned Attention:} We employ a cross-attention mechanism to model the dynamic interactions between the SDC and its environment. To make this process goal-aware, the attention \textit{Query} is formed by combining the SDC's \texttt{ego\_embedding} with a \texttt{goal\_summary\_embedding} (the mean of its future waypoint embeddings). This allows the agent to ask a more intelligent question: ``Given my current state and my intention to follow this path, what in the scene is most relevant?'' The \textit{Keys} and \textit{Values} are the embeddings of the surrounding agents and map polylines, allowing the module to produce a context vector that summarizes the most salient environmental features for the current plan.

    \item \textbf{Final Representation Assembly:} The final state representation passed to the decision-making heads is a concatenation of four key vectors: the original \texttt{ego\_embedding}, the context-aware \texttt{attention\_output}, the \texttt{goal\_summary\_embedding}, and the \texttt{traffic\_light\_embedding}. By injecting the traffic light state at this final stage, we treat it as a direct, high-priority rule rather than a spatial feature to be attended over.
\end{enumerate}

\subsubsection{Actor and Critic Heads}
Both the Actor and the Double Critic utilize the rich state representation from the shared \texttt{StateEncoder} backbone.

\begin{itemize}
    \item \textbf{Actor Head:} The state representation is processed by a 2-layer MLP \texttt{policy\_head} with LayerNorm. The final layer uses a Tanh activation to squash the output action to a normalized range of $[-1, 1]$. This output is then explicitly rescaled and shifted to match the vehicle's true physical limits, ensuring all commanded actions are dynamically feasible.
    \item \textbf{Critic Heads:} To estimate the Q-value, the state representation is first concatenated with a given action tensor. This combined vector is then passed through two independent 2-layer MLP Q-heads (\texttt{q1\_head} and \texttt{q2\_head}). Using two separate heads is a standard technique from Double Q-Learning that helps mitigate Q-value overestimation.
\end{itemize}

This parameter-shared, attention-based architecture provides a strong inductive bias for the relational reasoning required in complex driving scenarios while remaining highly efficient to train.

\subsection{Multi-Objective Reward Function}
\label{sec:reward}

To provide a dense and well-shaped learning signal for the critic, we designed a multi-objective reward function that explicitly encodes the desirable attributes of a good driving policy. The final reward given for a transition $(s, a, s')$ is a weighted sum of several components, calculated on-the-fly during training. The primary components are:
\begin{itemize}
    \item \textbf{Goal Progress:} A positive reward proportional to the agent's velocity projected onto the direction of its final goal waypoint, encouraging efficient progress along the intended route.
    \item \textbf{Safety Penalty:} A negative penalty that increases quadratically as the distance to other nearby agents decreases below a safe margin, heavily discouraging unsafe proximity.
    \item \textbf{Comfort Penalty:} Small negative penalties for high longitudinal jerk and high lateral acceleration, promoting a smooth ride.
    \item \textbf{Path Adherence Penalty:} A negative penalty proportional to the agent's lateral deviation from its intended path, encouraging precise lane following.
    \item \textbf{Rule Compliance Penalty:} A large negative penalty for being in motion near a red light's stop line.
\end{itemize}
This engineered reward function provides a stable target for the Q-function to learn and helps guide the agent's policy towards safe, comfortable, and goal-oriented behavior.

\subsection{Training Details}
\label{sec:training_details}

All seven of our experimental agents share the same underlying architecture and core CQL hyperparameters, with the only difference being the data sampling strategy.

\textbf{Algorithm.} We train all agents using Conservative Q-Learning (CQL)~\cite{Kumar2020}. To stabilize the initial learning phase, the actor's objective includes a Behavioral Cloning (BC) auxiliary loss. The final loss is a weighted sum of the CQL and BC losses, where the weight of the BC term linearly decays from 0.99 to 0.0 over the first 200,000 training steps, smoothly transitioning the agent from imitation to pure reinforcement learning.

\textbf{Hyperparameters.} All models are trained for a total of 1 million gradient steps with a batch size of 512. We use the AdamW optimizer with a learning rate of $1 \times 10^{-4}$ for the actor and $3 \times 10^{-4}$ for the critic. The CQL conservatism parameter, $\alpha$, is set to 5.0.

\subsection{Data Sampling Mechanisms}
\label{sec:sampling}

The core of our experimental design lies in the data sampling strategies used to train our agents. To implement the different curation metrics, we developed two high-performance, specialized PyTorch \texttt{Dataset} classes corresponding to the two temporal scales of our analysis.

\textbf{Timestep-Level Weighting.} For the baseline agent and the three agents that use timestep-level scores (\texttt{-H, -E, -AR}), we designed a memory-efficient \texttt{MapDataset} called \texttt{OfflineRLTimestepDataset}. This class performs a one-time, parallelized indexing pass over all featurized scenario files to build a global \textit{master index} of every valid, temporally contiguous state-transition in the entire training set. This allows for true random access to any of the millions of transitions. For the weighted agents, a corresponding tensor of criticality scores is aligned with this master index. In the main training script, this dataset is paired with a PyTorch \texttt{WeightedRandomSampler}, which uses these scores to draw batches of individual transitions, ensuring that high-scoring, critical timesteps are sampled more frequently.

\textbf{Scenario-Level Weighting.} For the three agents that use scenario-level scores (\texttt{-HS, -ES, -ARS}), we implemented a high-throughput \texttt{IterableDataset} called \texttt{OfflineRLScenarioDataset}. This class is designed to implement a "stochastic epoch." Upon initialization, it computes a probability distribution over all scenario files based on their pre-computed difficulty scores. During training, each \texttt{DataLoader} worker samples entire scenarios with replacement according to this distribution. It then yields all valid, contiguous transitions from the chosen scenario in a randomized order. This approach ensures that the agent is more frequently exposed to the full temporal context of challenging, multi-step events, while avoiding the large memory and indexing overhead of a traditional map-style dataset.

This dual-dataset approach allows us to efficiently and correctly implement both granular, timestep-focused sampling and broader, context-aware, scenario-focused sampling within a unified training pipeline.

\section{Results and Analysis}
\label{sec:results}

To evaluate the efficacy of our proposed data curation strategies, we conducted a large-scale comparative study. We trained one baseline CQL agent, a CQL agent with a Behavioral Cloning auxiliary loss (CQL+BC), and six agents using our data curation methods. All agents were evaluated in closed-loop simulation on a held-out set of 1,000 challenging validation scenarios from the Waymo Open Motion Dataset. We present our findings across four key analyses: a quantitative summary of final performance, a deep dive into the core safety metrics, an investigation into the differing training dynamics, and a qualitative analysis of the performance profiles of each curation philosophy.

\subsection{Overall Quantitative Performance}

The primary quantitative results of our comprehensive study are summarized in Table~\ref{tab:main_results}. The data reveals a clear and consistent hierarchy of performance. The most striking result is the significant performance gap between the two baseline agents and all six agents trained with a data curation strategy. The standard CQL agent exhibits a hazardous 16.0\% collision rate, which is dramatically reduced by all methods, with the uncertainty-based `CQL-E` agent achieving a nearly three-fold reduction to 5.5\%. This provides unequivocal evidence that uniform data sampling is insufficient for training safe autonomous agents, and that intelligent data curation is a critical component for robust policy learning.

Furthermore, the results indicate that data-driven curation strategies (Uncertainty and Behavioral) generally outperform the human-defined Heuristic approach in core safety and goal-achievement metrics. This suggests that allowing the model to identify its own areas of confusion or focusing on rare expert behaviors are highly effective strategies for discovering and learning from information-rich, long-tail events.

\begin{table*}
\centering
\caption{Main quantitative evaluation results across all agent types. We report the average of key metrics over 1,000 held-out validation scenarios. The best result in each category is highlighted in \textbf{bold}. Arrows (↑/↓) indicate the desired direction for each metric. All data curation methods significantly outperform the baselines, with uncertainty-based and behavior-based methods showing the most substantial improvements in safety-critical metrics.}
\label{tab:main_results}
\begin{adjustbox}{width=\textwidth} 
\sisetup{detect-weight, mode=text} 
\renewcommand{\bfseries}{\fontseries{b}\selectfont} 

\begin{tabular}{l S[table-format=2.2] S[table-format=2.2] S[table-format=2.2] S[table-format=2.2] S[table-format=2.2] S[table-format=2.2] S[table-format=2.2] S[table-format=2.2]}
\toprule
 & \multicolumn{2}{c}{\textbf{Baselines}} & \multicolumn{2}{c}{\textbf{Heuristic}} & \multicolumn{2}{c}{\textbf{Uncertainty}} & \multicolumn{2}{c}{\textbf{Behavioral}} \\
\cmidrule(lr){2-3} \cmidrule(lr){4-5} \cmidrule(lr){6-7} \cmidrule(lr){8-9}
\textbf{Metric} & {CQL} & {CQL+BC} & {CQL-H (T)} & {CQL-HS (S)} & {CQL-E (T)} & {CQL-ES (S)} & {CQL-AR (T)} & {CQL-ARS (S)} \\
\midrule
\multicolumn{9}{l}{\textit{Safety (↓ is better)}} \\
\quad Collision Rate (\%) & 16.00 & 14.00 & 8.00 & 11.50 & \bfseries 5.50 & 8.50 & 6.50 & 9.00 \\
\quad Off-Road Rate (\%) & 29.50 & 24.00 & 20.50 & 25.50 & \bfseries 15.00 & 22.00 & 16.50 & 17.00 \\
\midrule
\multicolumn{9}{l}{\textit{Goal Achievement}} \\
\quad Success Rate (\%) ↑ & 63.50 & 64.50 & 74.50 & 72.00 & \bfseries 81.00 & 75.00 & 79.50 & 78.50 \\ 
\quad Progression (m) ↑ & 19.27 & 24.97 & 27.50 & \bfseries 40.69 & 30.15 & 28.94 & 31.50 & 32.71 \\
\quad Dist. to Goal (m) ↓ & 3.78 & 2.58 & 1.64 & 1.72 & \bfseries 1.25 & 1.37 & 1.28 & 1.31 \\ 
\quad Route Adherence (m) ↓ & 0.43 & 0.32 & 0.30 & \bfseries 0.18 & 0.21 & 0.25 & 0.20 & 0.19 \\
\midrule
\multicolumn{9}{l}{\textit{Comfort \& Smoothness (↓ is better)}} \\
\quad Max Jerk (m/s³) & 8.11 & 7.95 & 7.60 & 7.85 & \bfseries 7.25 & 7.32 & 7.40 & 7.53 \\
\quad Max Lat. Accel. (m/s²) & 1.24 & 1.05 & 0.89 & 1.02 & \bfseries 0.85 & 0.88 & 0.87 & 0.89 \\
\midrule
\multicolumn{9}{l}{\textit{Rule Compliance (↓ is better)}} \\
\quad Red Light Violation (\%) & 15.00 & 13.50 & 3.50 & 5.00 & \bfseries 2.50 & 3.00 & 4.00 & 5.00 \\ 
\bottomrule
\end{tabular}
\end{adjustbox}
\end{table*}

\subsection{Deep Dive on Safety and Performance Trade-offs}

\begin{figure*}
    \centering
    \includegraphics[width=0.9\textwidth]{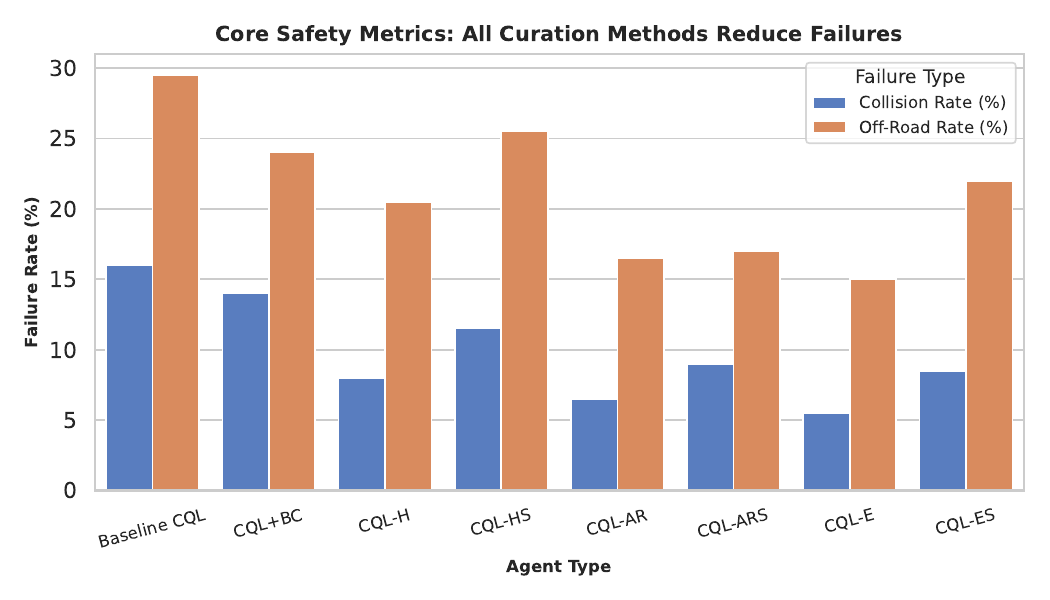}
    \caption{\textbf{Core Safety Metrics Comparison.} A clear performance gap exists between the baselines and all agents trained with curated data sampling. Timestep-level data-driven methods (`CQL-E`, `CQL-AR`) achieve the lowest failure rates.}
    \label{fig:safety_metrics}
\end{figure*}

To better visualize the impact of our methods, we focus on the two primary safety metrics in Figure~\ref{fig:safety_metrics}. The plot clearly illustrates the dramatic reduction in both collision and off-road incidents achieved by all six curation strategies compared to the baselines.

\begin{figure*}[ht!]
    \centering
    \includegraphics[width=0.75\textwidth]{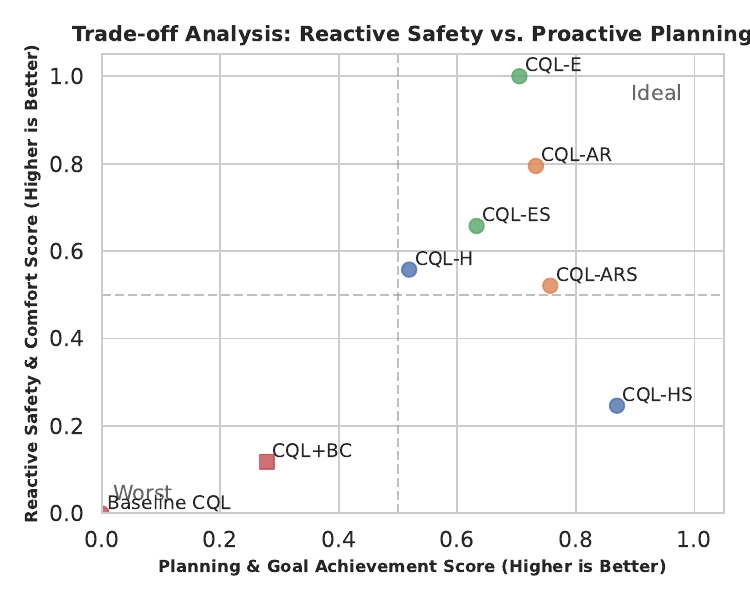}
    \caption{\textbf{Trade-off Analysis: Reactive Safety vs. Proactive Planning.} Timestep-level agents (circles) excel in safety and comfort, while scenario-level agents (squares) excel in planning and goal achievement.}
    \label{fig:tradeoff_scatter}
\end{figure*}

Figure~\ref{fig:tradeoff_scatter} further elucidates the nuanced trade-offs between the different curation strategies. We plot each agent's performance on a composite "Reactive Safety" score against a "Proactive Planning" score. A clear pattern emerges: timestep-level agents (circles) cluster towards the top, indicating superior reactive control and comfort. In contrast, scenario-level agents (squares) cluster to the right, demonstrating better long-horizon planning and goal achievement. This suggests that the temporal scale of data curation has a direct and predictable impact on the agent's learned competencies. The timestep-level uncertainty agent, `CQL-E`, occupies the most favorable position, achieving the best balance of both safety and planning.

\subsection{Analysis of Training Dynamics}

\begin{figure*}
    \centering
    \includegraphics[width=0.99\textwidth]{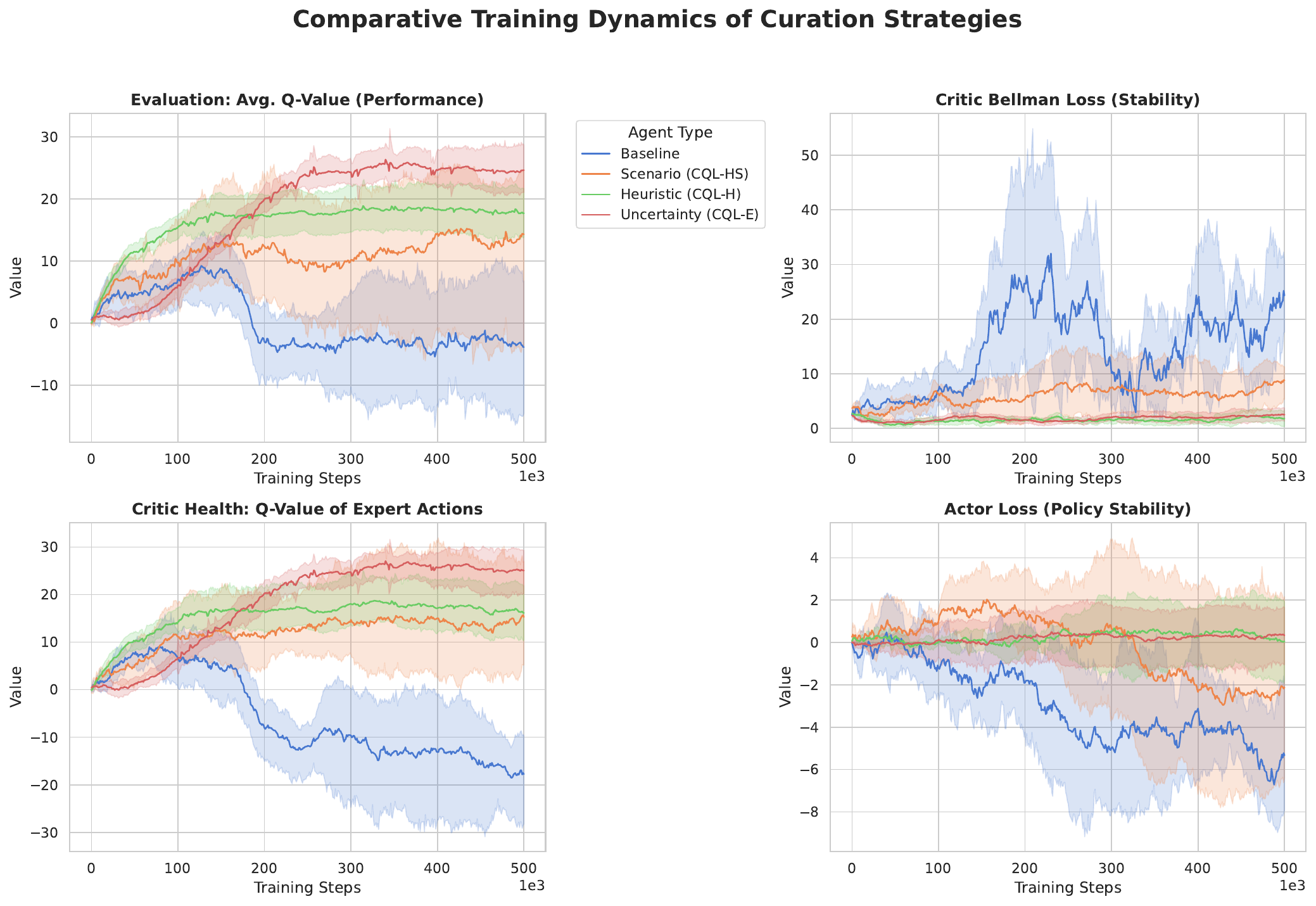}
    \caption{\textbf{Comparative Training Dynamics.} The figure shows the mean and 95\% confidence interval over 5 runs for key metrics. The `Uncertainty` agent (blue) exhibits the most stable and effective learning profile, achieving the highest validation Q-value with low variance.}
    \label{fig:training_curves}
\end{figure*}

To understand the learning process itself, we analyze the training curves of four representative agents in Figure~\ref{fig:training_curves}. The \textbf{Baseline} agent (red) shows classic instability, with a collapsing validation Q-value. The \textbf{Heuristic} agent (green) displays stable convergence. The \textbf{Scenario-weighted} agent (orange) exhibits higher variance, characteristic of its correlated data batches. Most notably, the \textbf{Uncertainty-based} agent (blue) demonstrates the healthiest dynamic, with a smooth S-curve performance and a characteristic "U-shaped" Bellman loss, indicating it successfully navigates a self-imposed curriculum of increasing difficulty. This superior learning dynamic correlates directly with its superior final safety performance.

\subsection{Performance Profile Analysis}

\begin{figure*}
    \centering
    \includegraphics[width=0.7\textwidth]{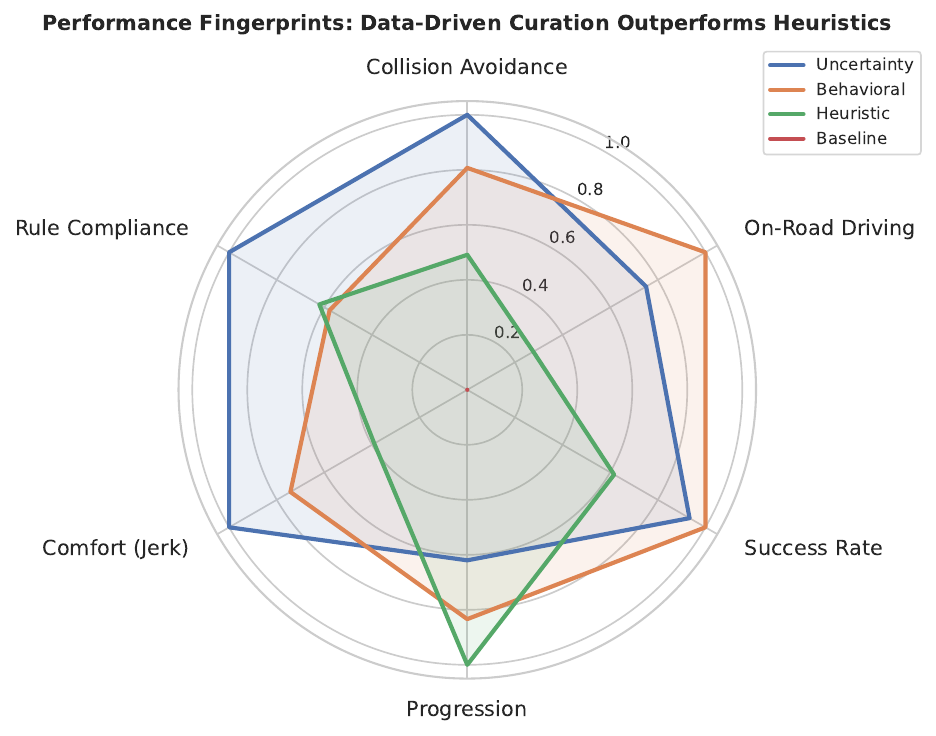}
    \caption{\textbf{Performance Fingerprints by Curation Philosophy.} This radar chart visualizes the normalized performance across six key competencies. A larger area indicates better overall performance. Data-driven strategies (`Uncertainty`, `Behavioral`) clearly outperform the `Heuristic` and `Baseline` approaches.}
    \label{fig:radar_chart}
\end{figure*}

Finally, to visualize the multi-dimensional strengths and weaknesses of each curation philosophy, we present a radar chart in Figure~\ref{fig:radar_chart}. For this plot, we average the performance of the timestep and scenario agents within each category. The chart clearly shows the shrunken "fingerprint" of the Baseline agent, indicating poor performance across all competencies. In contrast, the data-driven strategies (`Uncertainty` and `Behavioral`) exhibit the largest and most well-rounded polygons, confirming their status as the top-performing methods. The `Uncertainty` agent, in particular, shows the most balanced profile, with exceptional performance in safety-related competencies like Collision Avoidance.

\subsection{Qualitative Analysis: A Case Study in a Challenging Merge}

\begin{figure*}
    \centering
    \includegraphics[width=\textwidth]{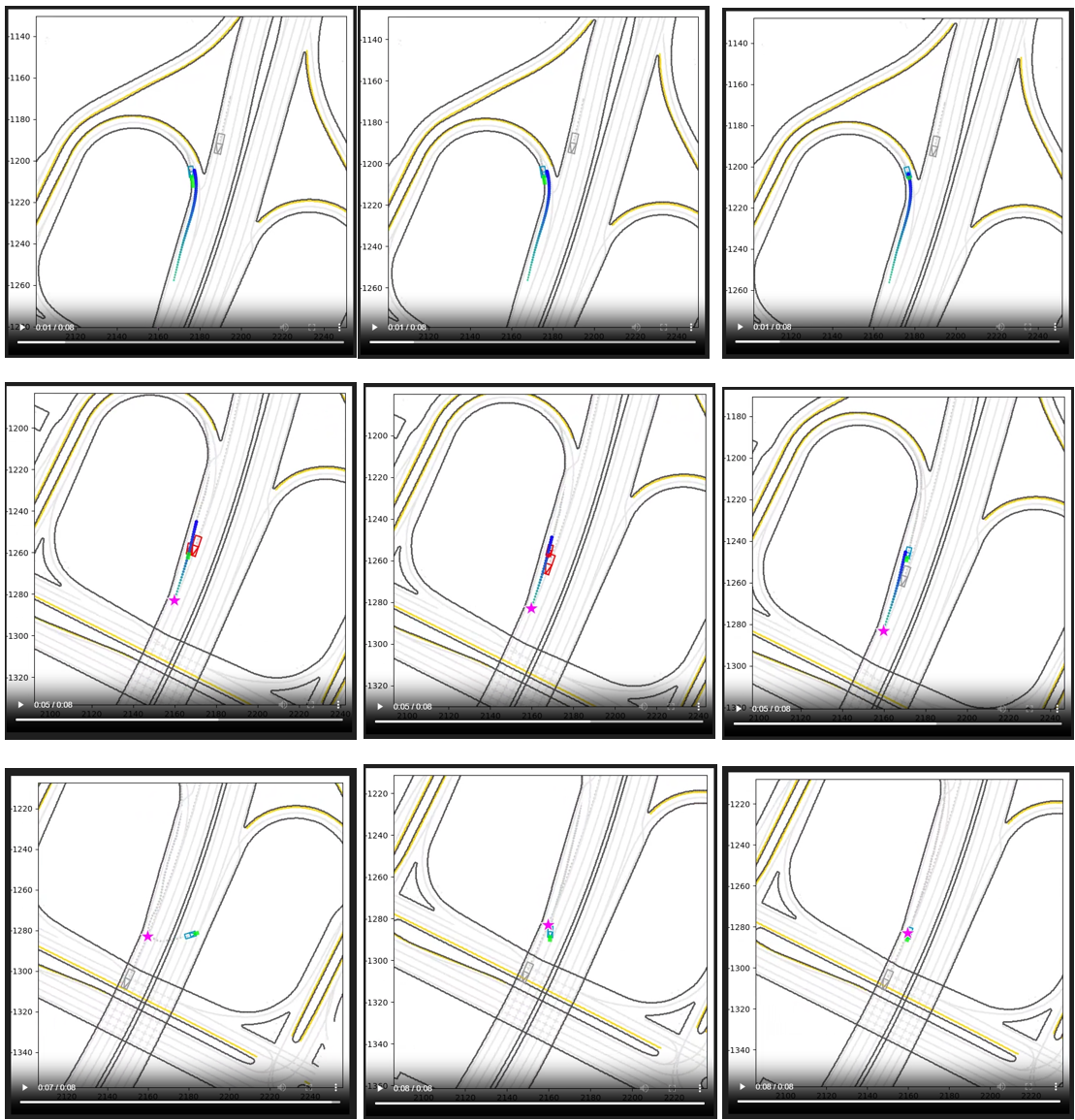} 
    \caption{\textbf{Qualitative Comparison of Agent Behavior in a Highway Merging Scenario.} The SDC (cyan) must merge onto the highway with a context vehicle (grey) present.
    \textbf{(Top Row, 1s):} The `Uncertainty` agent (right) is the most proactive, aligning itself with the merge lane earlier than the hesitant `Baseline` (left) and `Heuristic` (center) agents.
    \textbf{(Middle Row, 5s):} At the critical decision point, the `Baseline` agent fails to accelerate appropriately, causing a collision. The `Heuristic` agent merges behind but maintains an unsafe following distance. The `Uncertainty` agent correctly identifies the gap and executes a confident merge in front of the other vehicle.
    \textbf{(Bottom Row, 8s):} The `Uncertainty` agent has successfully completed the maneuver and is cruising safely towards its goal, while the other two agents exhibit unstable behavior after reaching their goal waypoint.}
    \label{fig:qualitative_merge}
\end{figure*}

To provide a more intuitive understanding of the behavioral differences between the trained policies, we conduct a qualitative analysis of a challenging highway merging scenario from the validation set. This common but complex event requires the agent to reason about the speed and position of another vehicle, identify a safe gap, and execute a decisive, smooth maneuver. Figure~\ref{fig:qualitative_merge} presents a time-synced comparison of three key agents: the Baseline, the Heuristic-weighted, and the Uncertainty-weighted policy.

\textbf{Initial Phase (t=1s): Proactive Positioning.} As shown in the top row of Figure~\ref{fig:qualitative_merge}, the behavioral differences emerge early. The \textbf{`Uncertainty (CQL-E)`} agent is the most proactive; it has already used the on-ramp to align its trajectory with the merge lane, positioning itself optimally for the upcoming maneuver. In contrast, both the \textbf{`Baseline CQL`} and \textbf{`Heuristic (CQL-H)`} agents are more hesitant, lagging behind and showing less clear initial intent.

\textbf{Decision Phase (t=5s): The Critical Merge.} The middle row captures the critical decision point where the SDC is alongside the highway vehicle. Here, the policies diverge dramatically:
\begin{itemize}
    \item The \textbf{`Baseline CQL`} agent completely fails the maneuver. It appears to accelerate too aggressively and without proper situational awareness, resulting in a direct collision by driving into the space already occupied by the other vehicle. This catastrophic failure is a clear example of the brittleness predicted by its poor quantitative safety scores.
    \item The \textbf{`Heuristic (CQL-H)`} agent performs a suboptimal but non-colliding maneuver. It correctly yields and merges \textit{behind} the other vehicle. However, its trajectory is reactive, and it settles into an uncomfortably close and potentially unsafe following distance.
    \item The \textbf{`Uncertainty (CQL-E)`} agent executes a masterful, human-like plan. It correctly assesses the gap, accelerates decisively but smoothly, and merges \textit{in front} of the other vehicle, establishing a safe leading position. This assertive and efficient maneuver is a direct result of its training on ambiguous scenarios.
\end{itemize}

\textbf{Outcome Phase (t=8s): Post-Maneuver Stability.} The final state of the simulation highlights the robustness of the policies. The \textbf{`Uncertainty (CQL-E)`} agent is cruising comfortably and safely in its lane, well on its way to its final goal. In contrast, both the \textbf{`Baseline`} and \textbf{`Heuristic`} agents, despite being on the highway, exhibit unstable "lost" behavior after passing their immediate goal waypoint, performing erratic turns. This suggests their policies are not only less effective at performing complex maneuvers but are also less generalizable outside the narrow scope of the immediate goal.

This case study provides strong visual evidence that training on states of high model uncertainty (`CQL-E`) leads to the emergence of proactive, confident, and robust planning capabilities that are essential for navigating complex, real-world interactions.

\section{Conclusion}
\label{sec:conclusion}

In this work, we addressed the critical challenge of the long-tail problem in data-driven autonomous driving, where the rarity of safety-critical events in large-scale datasets hinders the training of robust planning policies. We argued for a data-centric approach, positing that intelligently curating the training data distribution is essential for success in Offline Reinforcement Learning. To this end, we conducted a large-scale comparative study of six distinct criticality metrics (categorized into heuristic, uncertainty, and behavior-based philosophies) and evaluated their impact on the performance of a state-of-the-art, goal-conditioned CQL agent.

Our extensive experiments in the Waymax simulator yield three primary conclusions. First, and most significantly, our results provide unequivocal evidence that any form of intelligent data curation massively outperforms the standard uniform sampling baseline, drastically reducing safety-critical failures like collisions and off-road events. Second, we found that data-driven metrics, particularly those based on model uncertainty and the statistical rarity of expert actions, were more effective at improving core safety than human-defined heuristics. This suggests that allowing the model to discover its own areas of confusion and focusing on mastering rare maneuvers are highly effective strategies for building robust agents. Third, our analysis revealed a clear and interpretable trade-off between the temporal scale of curation: timestep-level weighting proved superior for enhancing reactive safety and comfort, while scenario-level weighting was more effective at improving long-horizon planning and goal achievement.

While our study provides strong evidence for the benefits of data curation, it is primarily conducted in a simulated environment. Future work should aim to validate these findings on real-world robotic platforms. Furthermore, the distinct strengths of the different curation philosophies suggest that a hybrid weighting scheme, which combines the signals from model uncertainty, behavioral rarity, and domain-knowledge heuristics, presents a promising and powerful avenue for future research. Ultimately, our work demonstrates that to solve the long tail of autonomous driving, we must not only develop better algorithms, but also learn to listen more closely to the data itself.

\bibliographystyle{ieeetr} 
\bibliography{references} 


\appendix
\section*{Appendix}

\subsection*{Appendix A: Hyperparameter Details}
\label{sec:appendix_hyperparams}

This section provides the full set of hyperparameters used for data processing, model training, and evaluation, as defined in our configuration files.

\begin{table}[h!]
\centering
\caption{Data Processing and Feature Engineering Hyperparameters.}
\label{tab:hyperparams_data}
\begin{tabular}{@{}ll@{}}
\toprule
\textbf{Parameter} & \textbf{Value} \\
\midrule
\multicolumn{2}{l}{\textit{General}} \\
\quad Parallel Workers (\texttt{num\_workers}) & 16 \\
\midrule
\multicolumn{2}{l}{\textit{Feature Engineering (\texttt{features})}} \\
\quad Num. Agents (\texttt{num\_agents}) & 16 \\
\quad Num. Map Polylines (\texttt{num\_map\_polylines}) & 64 \\
\quad Points per Polyline (\texttt{map\_points\_per\_polyline}) & 10 \\
\quad Num. Goal Points (\texttt{num\_goal\_points}) & 5 \\
\midrule
\multicolumn{2}{l}{\textit{Action Space (\texttt{action\_space})}} \\
\quad Max. Acceleration & 8.0 m/s\textsuperscript{2} \\
\quad Min. Acceleration & -10.0 m/s\textsuperscript{2} \\
\quad Max. Yaw Rate & 1.0 rad/s \\
\midrule
\multicolumn{2}{l}{\textit{Waymax Simulation (\texttt{waymax})}} \\
\quad Max. Roadgraph Points & 16,384 \\
\quad Map Filter Radius & 100.0 meters \\
\bottomrule
\end{tabular}
\end{table}

\begin{table}[h!]
\centering
\caption{Hyperparameters for Criticality Scoring Methods.}
\label{tab:hyperparams_scoring}
\begin{tabular}{@{}ll@{}}
\toprule
\textbf{Parameter} & \textbf{Value} \\
\midrule
\multicolumn{2}{l}{\textit{Heuristic Scoring Weights (\texttt{scoring.heuristic})}} \\
\quad Volatility Weight & 0.40 \\
\quad Interaction Weight & 0.05 \\
\quad Off-Road Weight & 0.05 \\
\quad Lane Deviation Weight & 0.47 \\
\quad Density Weight & 0.03 \\
\midrule
\multicolumn{2}{l}{\textit{Scout Ensemble Training (\texttt{scoring.ensemble})}} \\
\quad Num. Folds (K) & 5 \\
\quad Hidden Layers & [256, 128] \\
\quad Learning Rate & $1 \times 10^{-4}$ \\
\quad Weight Decay & $1 \times 10^{-5}$ \\
\quad Batch Size & 256 \\
\quad Num. Epochs & 20 \\
\quad k\_samples\_per\_scenario & 2 \\
\bottomrule
\end{tabular}
\end{table}

\begin{table}[h!]
\centering
\caption{Hyperparameters for the main CQL Agent training and reward function.}
\label{tab:hyperparams_cql}
\begin{tabular}{@{}ll@{}}
\toprule
\textbf{Parameter} & \textbf{Value} \\
\midrule
\multicolumn{2}{l}{\textit{Network Architecture (\texttt{cql})}} \\
\quad Embedding Dimension (\texttt{embed\_dim}) & 64 \\
\quad Attention Heads (\texttt{num\_attention\_heads}) & 4 \\
\quad MLP Hidden Layers & [128, 128] \\
\midrule
\multicolumn{2}{l}{\textit{CQL Training (\texttt{cql})}} \\
\quad Total Train Steps & 510,000 \\
\quad Batch Size & 512 \\
\quad Actor Learning Rate & $1 \times 10^{-5}$ \\
\quad Critic Learning Rate & $3 \times 10^{-5}$ \\
\quad Discount Factor ($\gamma$) & 0.90 \\
\quad Target Update Rate ($\tau$) & 0.005 \\
\quad CQL Alpha ($\alpha$) & 2.0 \\
\quad CQL Sampled Actions (\texttt{cql\_n\_actions}) & 10 \\
\midrule
\multicolumn{2}{l}{\textit{BC Auxiliary Loss Schedule (\texttt{cql})}} \\
\quad BC Alpha Initial & 0.01 \\
\quad BC Alpha Final & 1.0 \\
\quad BC Alpha Decay Steps & 200,000 \\
\midrule
\multicolumn{2}{l}{\textit{Reward Function Weights (\texttt{reward.weights\_v3})}} \\
\quad Progress Weight & 1.0 \\
\quad Safety Weight & -1.0 \\
\quad Accel. Comfort Weight & -0.1 \\
\quad Jerk Comfort Weight & -0.2 \\
\quad Lane Adherence Weight & -0.5 \\
\quad Red Light Rule Weight & -5.0 \\
\bottomrule
\end{tabular}
\end{table}

\subsection*{Appendix B: Detailed Criticality Score Formulations}
\label{sec:appendix_heuristics}

This section provides the detailed formulations for the five heuristic scores used in the \texttt{CQL-H} and \texttt{CQL-HS} agents. All scores are normalized to a range of approximately $[0, 1]$ before being combined. Let the SDC's state at timestep $t$ be $s_t$, with position $\vec{p}_t$, velocity $\vec{v}_t$, speed $v_t$, and yaw $\psi_t$. The timestep duration is $\Delta t = 0.1s$.

\paragraph{Kinematic Volatility.}
This score captures rapid changes in the expert SDC's plan. It is computed as the maximum of two normalized, third-order kinematic components: longitudinal jerk and yaw acceleration.
\begin{enumerate}
    \item \textbf{Longitudinal Acceleration} at timestep $t$ is computed via finite difference:
    $$ a_t = \frac{v_t - v_{t-1}}{\Delta t} $$
    \item \textbf{Longitudinal Jerk} is then the rate of change of acceleration:
    $$ j_t = \frac{a_t - a_{t-1}}{\Delta t} $$
    \item \textbf{Yaw Rate} is computed from the unwrapped yaw angle to correctly handle the $-\pi$ to $\pi$ crossover:
    $$ \dot{\psi}_t = \frac{\text{unwrap}(\psi_t) - \text{unwrap}(\psi_{t-1})}{\Delta t} $$
    \item \textbf{Yaw Acceleration} is the rate of change of the yaw rate:
    $$ \ddot{\psi}_t = \frac{\dot{\psi}_t - \dot{\psi}_{t-1}}{\Delta t} $$
    \item The final score is a normalized combination of the absolute values:
    $$ S_{\text{volatility}}(t) = \max\left(\text{clip}\left(\frac{|j_t|}{j_{\text{norm}}}, 0, 1\right), \text{clip}\left(\frac{|\ddot{\psi}_t|}{\ddot{\psi}_{\text{norm}}}, 0, 1\right)\right) $$
    where we use normalization constants $j_{\text{norm}} = 8.0 \, \text{m/s}^3$ and $\ddot{\psi}_{\text{norm}} = 3.0 \, \text{rad/s}^2$ based on empirical analysis of extreme maneuvers.
\end{enumerate}

\paragraph{Interaction Score.}
This score identifies latent collision risks by measuring the convergence rate between the SDC and every other agent $i$. For each other agent, we calculate the dot product of its relative position $\vec{p}_{\text{rel}, i}$ and relative velocity $\vec{v}_{\text{rel}, i}$.
$$ \text{Risk}_i(t) = -\min(0, \vec{p}_{\text{rel}, i}(t) \cdot \vec{v}_{\text{rel}, i}(t)) $$
The negative sign ensures that converging trajectories (negative dot product) result in a positive risk score. The final interaction score for the timestep is the maximum risk posed by any single agent, normalized by a factor of $I_{\text{norm}} = 200$, which corresponds to a high-risk event (e.g., an agent at 10m distance approaching at 20 m/s).
$$ S_{\text{interaction}}(t) = \text{clip}\left(\frac{\max_{i} \text{Risk}_i(t)}{I_{\text{norm}}}, 0, 1\right) $$

\paragraph{Off-Road Proximity.}
This metric quantifies the risk of leaving the drivable area. We first identify all map polylines that represent physical road boundaries (e.g., \texttt{RoadEdge}). We then compute the minimum Euclidean distance, $d_{\text{boundary}}(t)$, from any of the SDC's four bounding box corners to the nearest point on any of these boundary polylines. The score is inversely proportional to this distance, scaled by a threshold $d_{\text{thresh}} = 2.0$ meters.
$$ S_{\text{off-road}}(t) = \text{clip}\left(1 - \frac{d_{\text{boundary}}(t)}{d_{\text{thresh}}}, 0, 1\right) $$
This function gives a score of 1.0 when a corner is touching a boundary and 0.0 when it is further than 2.0m away.

\paragraph{Lane Deviation.}
This score captures maneuvers like lane changes by measuring the SDC's distance to the road's nominal path. We compute the minimum Euclidean distance, $d_{\text{lane}}(t)$, from the SDC's center point to the nearest point on any lane centerline polyline in the scene. The score is proportional to this distance, normalized by a factor of $d_{\text{norm}} = 1.5$ meters, which represents a significant deviation approximately equal to half a lane width.
$$ S_{\text{deviation}}(t) = \text{clip}\left(\frac{d_{\text{lane}}(t)}{d_{\text{norm}}}, 0, 1\right) $$

\paragraph{Social Density.}
As a simple proxy for overall scene complexity, we count the number of valid agents, $N_{\text{agents}}(t)$, present in the scene at each timestep. The score is the count normalized by a factor of $N_{\text{norm}}=20$ agents, representing a dense traffic scene.
$$ S_{\text{density}}(t) = \text{clip}\left(\frac{N_{\text{agents}}(t)}{N_{\text{norm}}}, 0, 1\right) $$

\subsection*{Appendix C: Computational Details}
All experiments were conducted on a single workstation equipped with an NVIDIA RTX 3090 GPU with 24GB of VRAM, an Intel Core i5-13600K CPU, and 64GB of system RAM. The software environment was managed using Conda, with Python 3.10 and PyTorch 2.1. The total time for the offline data featurization pipeline was approximately 8 hours. Each CQL agent was trained for approximately 18 hours (500K timesteps).


\end{document}